\pdfoutput=1
\documentclass[11pt]{article}

\usepackage[final]{acl}

\usepackage{graphicx}
\usepackage{times}
\usepackage{latexsym}
\usepackage{booktabs}
\usepackage{amsmath}
\usepackage{multirow}
\usepackage{pgfplots}
\usepackage{algorithm}
\usepackage{algorithmic}

\usepackage[T1]{fontenc}
\usepackage[utf8]{inputenc}

\usepackage{microtype}

\usepackage{inconsolata}

\title{  \includegraphics[height=1em]{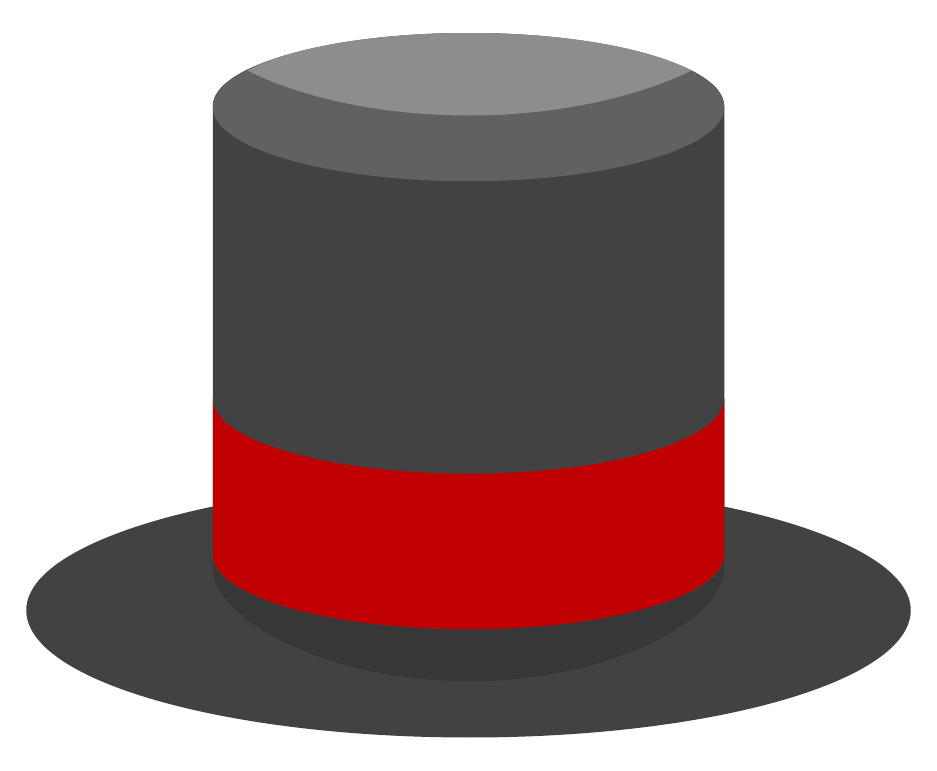} SirLLM: Streaming Infinite Retentive LLM}

\author{Yao Yao$^{1,2}$, Zuchao Li$^{3,*}$ and Hai Zhao$^{1,2,}$\thanks{$\ $  Corresponding author. This research was supported by the Joint Research Project of
Yangtze River Delta Science and Technology Innovation Community (No.
2022CSJGG1400), the National Natural Science Foundation of China (No. 62306216), the Natural Science Foundation of Hubei Province of China (No. 2023AFB816), the Fundamental Research Funds for the Central Universities (No. 2042023kf0133).} \\
$^{1}$Department of Computer Science and Engineering, Shanghai Jiao Tong University\\
$^{2}$MoE Key Lab of Artificial Intelligence, AI Institute, Shanghai Jiao Tong University\\
$^{3}$National Engineering Research Center for Multimedia Software, \\
School of Computer Science, Wuhan University, Wuhan, 430072, P. R. China \\
{\tt yaoyao27@sjtu.edu.cn, zcli-charlie@whu.edu.cn,}\\
{\tt zhaohai@cs.sjtu.edu.cn}\\
}

\begin{document}
\maketitle
\begin{abstract}
 % we propose the Streaming infinite retentive LLM (SirLLM) in this paper. Utilizing the newly introduced metric of token entropy to calculate the criticality of each token, we selectively preserve key tokens in the kv cache. This approach significantly enhances the model's memory capabilities in the context of infinitely long streaming dialogues.

 % To validate the effectiveness of SirLLM, we conducted experiments on three datasets: grocery shopping, DailyDialog, and rock-paper-scissors. The results of these experiments effectively demonstrate the enhanced memory capabilities of SirLLM.
 
% As Large Language Models (LLMs) find extensive applications across various fields, the ability to process infinite input lengths and possess a certain degree of memory capability becomes a necessity. However, studies have shown that when inputs exceed the pre-trained text length of LLMs, there is a dramatically decline in text generation capabilities. Additionally, current methods that allow LLMs to handle infinite streaming dialogue inputs significantly impair the model's long-term memory capabilities. 

As Large Language Models (LLMs) become increasingly prevalent in various domains, their ability to process inputs of any length and maintain a degree of memory becomes essential. However, the one-off input of overly long texts is limited, as studies have shown that when input lengths exceed the LLMs' pre-trained text length, there is a dramatic decline in text generation capabilities. Moreover, simply extending the length of pre-training texts is impractical due to the difficulty in obtaining long text data and the substantial memory consumption costs this would entail for LLMs. Recent efforts have employed streaming inputs to alleviate the pressure of excessively long text inputs, but this approach can significantly impair the model's long-term memory capabilities.

Motivated by this challenge, we introduce Streaming Infinite Retentive LLM (SirLLM), which allows LLMs to maintain longer memory during infinite-length dialogues without the need for fine-tuning. SirLLM utilizes the Token Entropy metric and a memory decay mechanism to filter key phrases, endowing LLMs with both long-lasting and flexible memory. We designed three distinct tasks and constructed three datasets to measure the effectiveness of SirLLM from various angles: (1) DailyDialog; (2) Grocery Shopping; (3) Rock-Paper-Scissors. Our experimental results robustly demonstrate that SirLLM can achieve stable and significant improvements across different LLMs and tasks, compellingly proving its effectiveness. When having a coversation, "A sir could forget himself," but SirLLM never does! Our code is publicly available at \href{https://github.com/Zoeyyao27/SirLLM}{https://github.com/Zoeyyao27/SirLLM}  
\end{abstract}

\section{Introduction}
%不限输入情况下的，更长记忆
% Recently, the popularity of large language models (LLM) has immediately skyrocketed, owing to the fact that natural language serves as an exceptionally intuitive interface which makes artificial intelligence accessible to a broad audience. The proliferation of LLMs, such as Llama ~\cite{touvron2023llama}, ChatGPT ~\cite{achiam2023gpt}, Mistral~\cite{jiang2023mistral}, has spurred the development of numerous NLP applications, including commonly seen tools such as chatbots \cite{bill2023fine,pandya2023automating}, writing assistants \cite{bhat2023approach}, programming assistants \cite{kazemitabaar2024codeaid}, and more. However, these applications often requires LLM posscess memory and 

% Recently, the popularity of large language models (LLMs) ~\cite{touvron2023llama,achiam2023gpt,jiang2023mistral}, has skyrocketed. 
% This surge is attributed to the fact that natural language acts as an incredibly intuitive interface, making artificial intelligence accessible to a wide audience. 
The proliferation of large language models (LLMs) ~\cite{touvron2023llama,achiam2023gpt,jiang2023mistral} has spurred the development of various NLP applications\cite{zhang2023arcgpt,yang2024batgpt,zhang2023generative,ma2024comprehensive,wang2024llm}, including widely-used tools like chatbots \cite{bill2023fine,pandya2023automating}, writing assistants \cite{bhat2023approach}, and programming assistants \cite{kazemitabaar2024codeaid}. These applications, aiming to enhance user interaction and conversational experience, often require infinite input length and a certain degree of memory capability. However, current LLMs are usually pre-trained on texts of limited length, and studies have shown that their text generation capabilities dramatically decline when input lengths exceed those of the pre-training texts \cite{xiao2023efficient,huang2023advancing}. Merely extending the length of pre-training texts is impractical, as acquiring infinitely long text data is exceedingly challenging, not to mention that it would result in substantial memory consumption for LLMs. Therefore, researching how to enable LLMs to handle infinite input lengths while maintaining memory capability is an urgent issue to be addressed.
\begin{figure*}[t]
    \centering
    \includegraphics[width=0.95\linewidth]{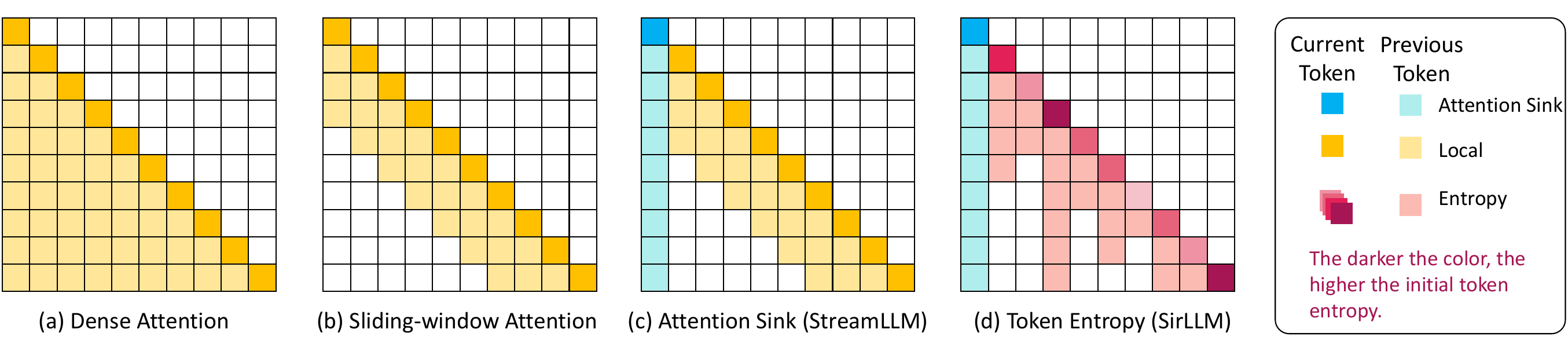}
    \caption{The visualization of SirLLM versus existing attention patterns.}
    \label{fig:attention_window}
\end{figure*}

\begin{figure*}[h]
    \centering
    \includegraphics[width=0.95\linewidth]{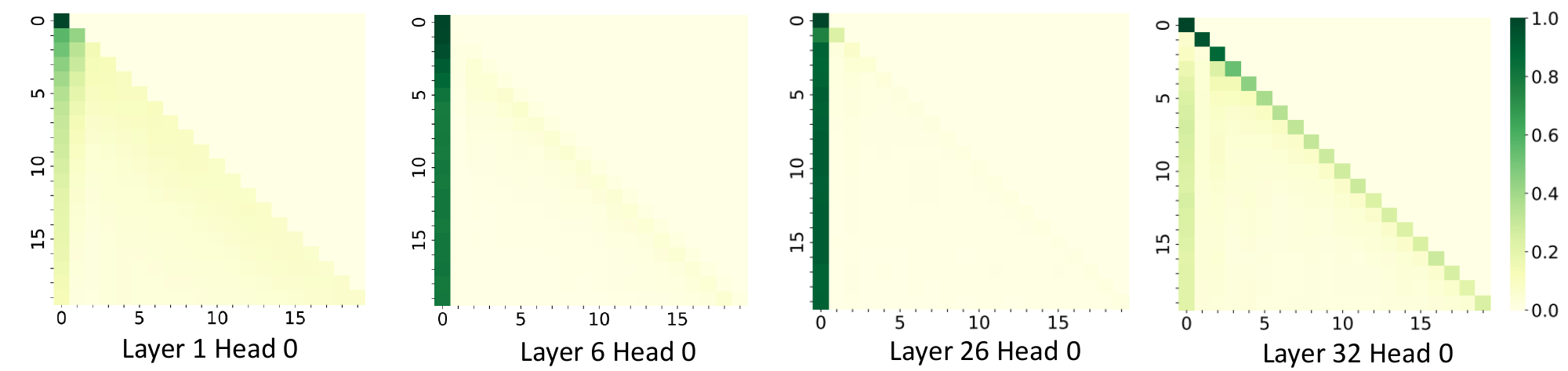}
    \caption{Attention sink phenomenon \cite{xiao2023efficient}. We visualize the average layer attention logits over 256 sentences, each with a length of 20, in Vicuna-7b-v1.3. We can see that in the shallow layers, a significant amount of the attention score is dedicated to the first tokens and in the final layer, the model focuses more on the recent tokens.}
    \label{fig:attention sink}
\end{figure*}

With the emergence of this demand, researchers have gradually shifted their focus towards exploring ways to expand the input context length of LLMs. A line of these studies has particularly focused on optimizing the attention mechanism of LLMs. 
\cite{beltagy2020longformer} first proposes the Sliding-window attention, as shown in Figure \ref{fig:attention_window} (a). By restricting each token to only attend to a certain number of recent tokens, this method reduces computational complexity. 
In deployment scenarios, LLMs utilize a Key-Value (KV) cache to store the key and value tensors of past tokens at each generation step to effectively reduces the need to recompute past key and value tensors, thereby significantly lowering computational overhead. 
Consequently, Sliding-window attention ensures a stable decoding speed even when the KV cache is full, thereby allowing for longer texts during the pre-training phase. However,  \citet{xiao2023efficient} discovered that this method does not truly achieve infinite input length, as the model's performance significantly deteriorates once the input length exceeds the size of the KV cache and intial tokens, however, receive a disproportionately higher amount of attention, a phenomenon termed as `attention sink`, as shown in Figure \ref{fig:attention sink}. Therefore, they proposed StreamLLM, as shown in Figure \ref{fig:attention_window} (b). StreamLLM enhances the potential of window attention by preserving the KV cache of the initial tokens, thereby achieving infinite length input in streaming conversations without finetuning. However, while Sliding-window Attention and StreamLLM ensure an expanded input length, each generated token only attends to recent tokens (and initial attention sink tokens), resulting in a loss of memory for earlier parts of the conversation. This leads to a significant forgetting issue in long-distance dialogues. Furthermore, as observed in Figure \ref{fig:attention sink}, the range of recent tokens that the model focuses on is not very extensive. This observation leads us to contemplate \textbf{whether it's possible for the model to concentrate only on key terms during a conversation, filtering out less important tokens}. By remembering only the crucial information, the model might be able to maintain a longer memory span in the context of infinitely long conversations. 
% More detailed related works please refer to Appendix \ref{append:related}.

% Therefore, in response to the aforementioned issues, we propose the Streaming infinite retentive LLM (SirLLM) in this paper. Utilizing the newly introduced metric of token entropy to calculate the criticality of each token, we selectively preserve key tokens in the kv cache. This approach significantly enhances the model's memory capabilities in the context of infinitely long streaming dialogues.

In response to the aforementioned challenges, we propose the Streaming Infinite Retentive LLM (SirLLM) in this paper, as illustrated in Figure \ref{fig:attention_window} (d). Initially, we employ an LLM to calculate the token entropy metric for each input token, thereby assessing their significance. Subsequently, tokens with higher token entropy values, deemed as key tokens, are preserved within the KV cache. This method enhances the model's memory capabilities in the context of infinitely long streaming dialogues. To validate the effectiveness of SirLLM, we conducted experiments across three distinct tasks: (1) DailyDialog: We created a multi-turn daily dialogue dataset based on the DailyDialog dataset \cite{li-etal-2017-dailydialog}. (2) Grocery Shopping: We developed a grocery shopping dataset. Users first inform the LLM about the groceries they need to purchase. Following this, users engage in multi-turn dialogues with the LLM, culminating in the users asking the LLM to recall the required groceries. (3) Rock-Paper-Scissors: We constructed a rock-paper-scissors dataset featuring three types of players, each with a preference for one of the three moves (rock, paper, scissors). Players engage in multiple rounds of rock-paper-scissors with the LLM, which is tasked with analyzing the user's historical preferences to maximize its winning rate. The results of these experiments effectively demonstrate the enhanced memory capabilities of SirLLM in infinite conversation.

\section{Related Work}
Many works \cite{DBLP:conf/nips/LiJXZCWY19,DBLP:conf/naacl/GuoAUONSY22,DBLP:journals/corr/abs-2308-16137,DBLP:conf/emnlp/AinslieOACFPRSW20,DBLP:journals/corr/abs-2309-12307} focused on expanding the input context length of LLMs by optimizing the attention mechanism. \citet{beltagy2020longformer} first proposes the sliding window attention, which let each token to only attend to a certain number of recent tokens. When the KV cache is full sliding window attention would discard the earliest token to preserve a stable decoding speed and performance. 
\citet{DBLP:journals/corr/abs-1904-10509} proposed the fixed Sparse Transformer. Formally, this method initially preserves the key and value states of recent tokens as local context information. Subsequently, it employs a column attention mechanism with a specified stride. This mechanism summarizes information from previous locations and propagates it to all future tokens, functioning as a form of global attention.
\citet{li2019enhancing} proposed a LogSparse self-attention where each element can only to attend to itself and its previous cells
with an exponential step size. 
% \citet{xiao2023efficient} introduced the attention sink phenomenon and proposed StreamLLM, a model designed to truly achieve infinite input length. During attention calculation, StreamLLM keeps the attention of the initial tokens and the recent tokens to achieve a stable performance under the infinite streaming conversation .
\citet{xiao2023efficient} introduced the attention sink phenomenon and proposed StreamLLM, a model specifically designed to achieve true infinite input length. StreamLLM, during its attention calculation, maintains the focus on both the initial tokens and the recent tokens. This approach ensures stable performance in the context of infinite streaming conversations. 

% \citet{DBLP:journals/corr/abs-2310-01801} introduced FastGen, an adaptive KV cache compression method for Large Language Models. FastGen employs four distinct compression strategies: Special Tokens, Punctuation, Locality, and Frequency. The method starts by analyzing the behavior of various attention heads, enabling it to select the most effective compression strategy for each head. When generating new tokens, FastGen optimizes KV cache management by applying the chosen compression strategy to each token, rather than simply appending new KV vectors. This approach leads to more efficient memory usage without compromising the model's performance.

However, the aforementioned approaches either save tokens with given stride, randomly select, or do not preserve the key-value (KV) cache of history tokens, leading to significant forgetting issues in the model. SirLLM addresses this by utilizing the LLM itself to calculate token entropy, selectively preserving the KV cache of tokens with the highest entropy. This method effectively conserves memory space, ensuring that only the most crucial information is retained.

Another line of related work is KV cache optimization \cite{DBLP:conf/nips/Zhang00CZC0TRBW23,DBLP:journals/corr/abs-2401-06104,DBLP:journals/corr/abs-2310-01801}. \citet{DBLP:journals/corr/abs-2310-01801} introduced FastGen, an adaptive KV cache compression method for Large Language Models. FastGen 
%implements four distinct compression strategies: Special Tokens, Punctuation, Locality, and Frequency. It 
begins by analyzing the behavior of various attention heads to select the most effective compression strategy for each and optimizes KV cache management when generating new tokens by applying the chosen compression strategy to each token, instead of merely appending new KV vectors.  \citet{DBLP:conf/nips/Zhang00CZC0TRBW23} proposed H$_2$O
, a KV cache eviction policy that dynamically balances recent tokens and heavy hitters. The eviction policy is framed as a dynamic submodular problem, using attention scores to retain the most influential tokens in the KV cache. A greedy algorithm provides theoretical guarantees for near-optimal performance. However, these works focus more on KV cache optimization rather than the streaming scenarios of multi-turn dialogues and enhancing the memory capabilities of LLMs.

Another category of work related to our research is context compression.  \citet{li-etal-2023-compressing} compress the input context by selecting the lexical units (tokens, phrases, sentences) with higher self-information computed by a base language model. \citet{DBLP:conf/emnlp/BerchanskyICDW23} proposed a token filtering method for optimizing retrieved long documents to speed up the decoding process. This method involves using mean cross-attention scores computed at a specific layer across all attention heads to eliminate less critical tokens. Then, only the top k\% of input tokens with the highest scores are retained and used in predicting subsequent tokens. 
Although retrieval-based methods can identify more accurate contexts based on input, they typically require greater computational and time resources. In contrast, SirLLM does not necessitate maintaining an additional vector database and does not disrupt the model's end-to-end computational process. SirLLM can significantly enhance the model's memory capabilities efficiently without modifying the model's architecture or requiring fine-tuning.
%focuses on proposing eliminating less critical tokens to speed up the decoding process. They

\section{Method}
\subsection{Preliminaries}
%介绍一下stream llm， 我们和stream llm 采用相同的position id 来进行position embedding

%attention window comparison figure
\citet{xiao2023efficient} proposed StreamLLM. They discovered that the model disproportionately focuses on initial tokens and break when removing initial tokens' KV cache. Therefore, based on the Sliding-window attention, instead of throwing away all of the previous KV cache except the recent token's KV cache, they keep the first initial tokens KV cache as shown in Figure \ref{fig:attention_window} (c). Figure \ref{fig:attention_window} (c) illustrates the StreamLLM process, which can be formulated as follows. We define the indices of the attention sink tokens and the recent tokens as $\mathrm{Id}_{sink}$ and $\mathrm{Id}_{recent}$, respectively: 
$$
\mathrm{Id}_{sink}=\{0,1,...,n_{sink}\}
$$
$$
\mathrm{Id}_{recent}=\{L-n_{recent}+1,...,l-1,l\}
$$
where, $n_{sink}$ and $n_{recent}$ denotes the KV cache size of the attention sink tokens and recent tokens respectively. $l$ denotes the total length of the past key-value states.

Then the StreamLLM only keeps the selected tokens' past key and value states:
$$\mathrm{KV}_{cache}=\mathrm{K}_{cache}[\mathrm{Id}_{sink},\mathrm{Id}_{recent}]$$
% $$\mathrm{V}_{cache}=\mathrm{V}_{cache}[\mathrm{Id}_{sink},\mathrm{Id}_{recent}]$$
where $\mathrm{X[Id]}$ indicates the selection of vectors from $\mathrm{X}$ using indices in using indices in $\mathrm{Id}$. 

% Another important trick StreamLLM used is that when determining the relative distance and injecting positional information, StreamLLM focuses on token positions within the cache, as opposed to their original positions in the text. For example, if the current cache contains tokens [0, 1, 2, 3, 5, 7, 11, 12] and is decoding the 13th token, the assigned positions are [0, 1, 2, 3, 4, 5, 6, 7] instead of the original text positions.

However, StreamLLM primarily focuses on recent tokens and the initial attention sink tokens. This raises an intriguing question: Could we conserve cache space occupied by recent tokens by only preserving the past key-value states of critical tokens? Such an approach would allow the model to access information from tokens over a longer time span, potentially enhancing its long-term memory and reducing the problem of forgetting. To address this issue, our first step is to define a metric that can accurately measure the importance of each token.

\subsection{Token Entropy}
Recent work \cite{li-etal-2023-compressing} has focused on context compression. This involves utilizing LLMs to calculate the information contained in each token, thereby compressing the input context to enhance the model's inference efficiency. Inspired by this, we use the token entropy metric to assess the significance of tokens. Given a input sequence $X=\{x_{1},x_{2},...,x_{n}\}$ , where $x_{i}$ denotes i-th token and $n$ denotes the total token number. We define the token entropy of the i-th token as:
$$
e_{i}=-\mathrm{logP}(x_{i}|x_{0},x_{1},...,x_{i-1})
$$
%The token with a higher token entropy means the token contains more information and hence is more critical.
A token with higher token entropy indicates that it contains more information and is therefore more critical. In our experiments, we utilize the LLM to calculate the generation probability of each token. This approach allows us to obtain the entropy of each token concurrently with its generation, without necessitating additional computational effort.

\begin{figure}
    \centering
    \includegraphics[width=1\linewidth]{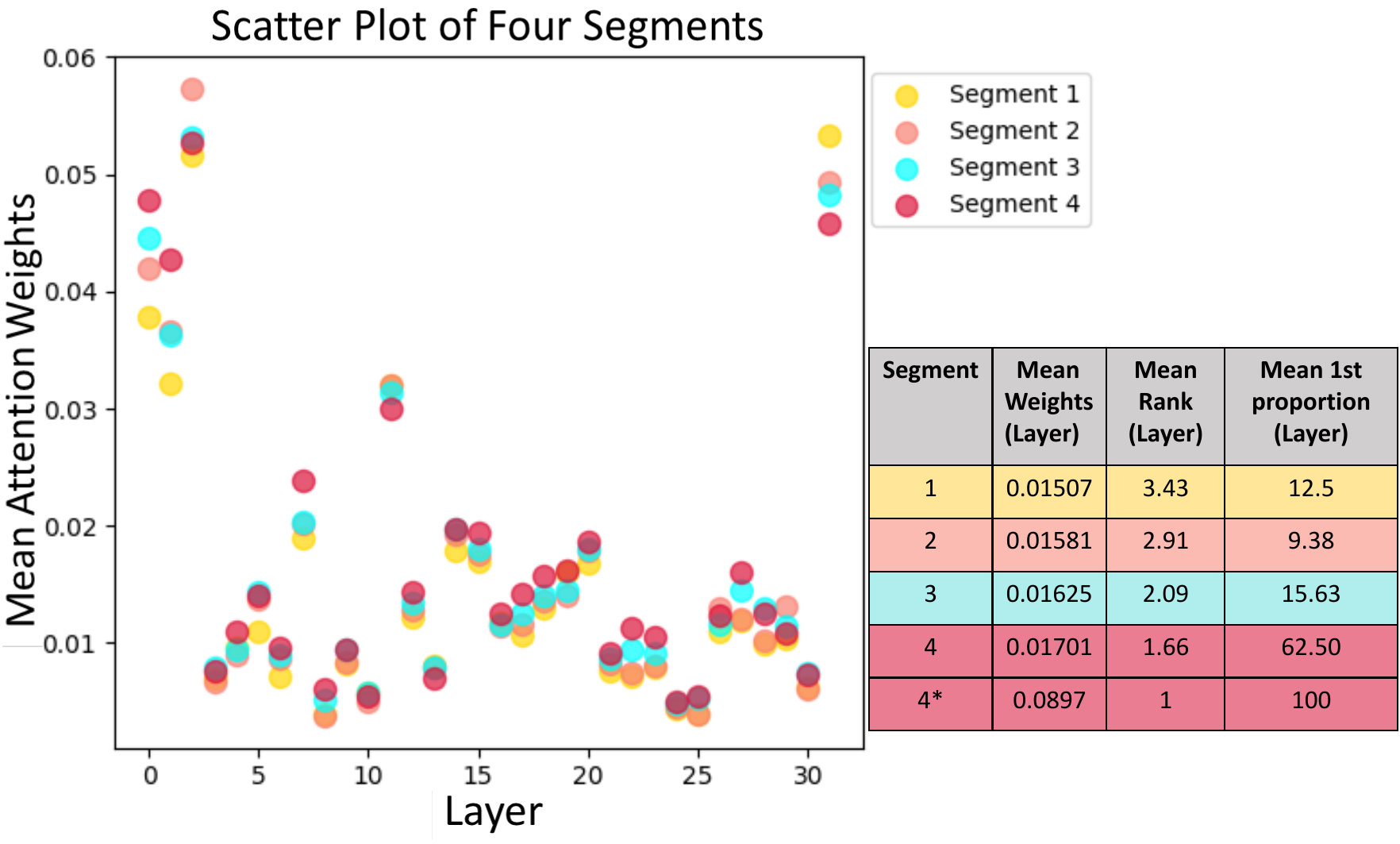}
    \caption{Scatter Plot of the average attention weights over 256 sentences at every layer. We divide the tokens into four segments based on token entropy, with segment 1 having the lowest entropy and segment 4 the highest. Mean Weights stands for the average attention weights across all layers. Mean Rank denotes the average ranking of each segment at every layer. Mean 1st proportion denotes the proportion of times each segment ranked first across all layers. The figure indicates that as token entropy increases, so does the attention that the LLM allocates to that token.}
    \label{fig:mean_attention_logtis}
\end{figure}

\paragraph{Does higher token entropy equate to increased LLM focus?}
To investigate whether tokens with higher entropy indeed carry more information and consequently garner more attention from LLMs, we extracted 256 sentences from the Wikitext corpus \cite{DBLP:conf/iclr/MerityX0S17}, focusing on the first 40 tokens of each sentence. To mitigate the attention sink effect, we omitted the first token, starting our analysis from the second token, thus providing a clearer view of the model's attention distribution across other tokens. The 40 tokens were divided into four segments based on token entropy, with segment 1 having the lowest entropy and segment 4 the highest. We calculated the average attention weights for each segment at every layer and plotted these values in a scatter plot, as shown in Figure \ref{fig:mean_attention_logtis}. For a more tangible understanding, we also computed the average attention weights across all layers for each segment. The results show that tokens in segments with higher entropy have higher attention weights. This pattern reinforces the hypothesis that higher entropy tokens, which are presumably less predictable and therefore more informative, are given priority by the LLM's attention mechanism.  This finding supports the validity of the token entropy metric as an indicator of a token's significance.

\subsection{ Streaming Infinite Retentive LLM}
%介绍利用self info来压缩kv cache
%framework overview figure
% 流程伪代码
\begin{figure*}
    \centering
    \includegraphics[width=1\linewidth]{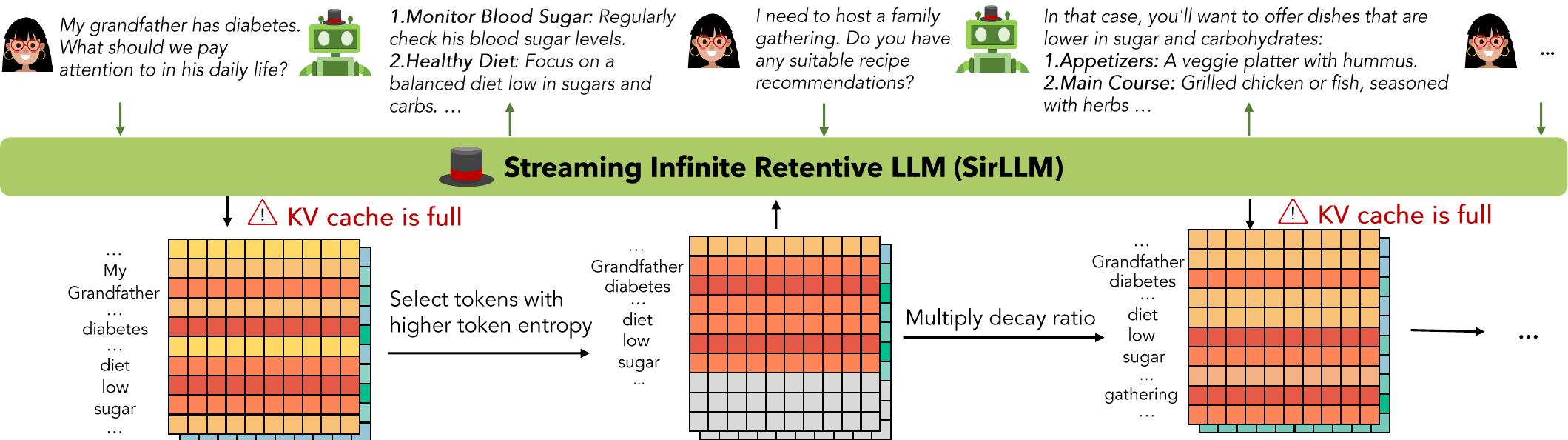}
    \caption{Framework overview of SirLLM. When the number of tokens stored in KV cache exceeds the pre-training length $L$, SirLLM calculates the entropy of each token and selects the tokens with the higher token entropy, thereby conserving space in the KV cache}
    \label{fig:sirllm_overview}
\end{figure*}
Upon obtaining the entropy values for each token, we enhance the model's memory capability by selectively preserving the key-value states of only the key tokens and propose Streaming Infinite Retentive LLM (SirLLM) as shown in Figure \ref{fig:sirllm_overview}. To elaborate further, we maintain both a key-value (KV) cache $\mathrm{KV}_{cache}$ and a token entropy cache $E$ in parallel. The token entropy cache stores the entropy values of tokens present in the KV cache.  When the number of tokens stored in $C$ exceeds the pre-training length $L$, we utilize $E$ to select the tokens with the higher token entropy, thereby conserving space in the KV cache:

$$E=\{e_1,e_2,...,e_l\}; \quad \mathrm{Id}_{entropy}=\mathrm{Top}_{k}(E)$$
% $$
% \mathrm{Id}_{entropy}=\mathrm{Top}_{k}(E)
% $$
$$\mathrm{KV}_{cache}=\mathrm{KV}_{cache}[\mathrm{Id}_{sink},\mathrm{Id}_{entropy}]$$
% $$\mathrm{V}_{cache}=\mathrm{V}_{cache}[\mathrm{Id}_{sink},\mathrm{Id}_{entropy}]$$
% $$C=C[\mathrm{K}_{cache},\mathrm{V}_{cache}]$$
$$E=E[\mathrm{Id}_{sink},\mathrm{Id}_{entropy}]$$
where $\mathrm{Top}_{k}$ denotes the selection of the top $k$ tokens with the highest token entropy. Higher token entropy implies a lower probability of the model generating the word, indicating such words carry more information and are likely to be key tokens.

Following StreamLLM, SirLLM concentrates on the token positions within the cache rather than their original positions in the text when determining relative distances and injecting positional information. For instance, if the current cache holds tokens [0, 1, 2, 3, 5, 7, 11, 12] and the model is in the process of decoding the 13th token, it assigns positions as [0, 1, 2, 3, 4, 5, 6, 7] instead of using the original text positions.
%is its approach to determining relative distances and injecting positional information. StreamLLM concentrates on the token positions within the cache rather than their original positions in the text. For instance, if the current cache holds tokens [0, 1, 2, 3, 5, 7, 11, 12] and the model is in the process of decoding the 13th token, it assigns positions as [0, 1, 2, 3, 4, 5, 6, 7] instead of using the original text positions.

However, simply preserving tokens with the highest token entropy, as previously described, can lead to a limitation in the KV cache. After lengthy multi-turn dialogues with users, the cache may become restricted to a few tokens with very high entropy, making it difficult for the cache to adapt. This could result in a 'rigid memory' within the model, lacking flexibility. An effective dialogue system should, like human memory, have a more flexible mechanism for long and short-term memory: the more distant the memory, the easier it is for the model to forget it. This approach ensures the freshness of the LLM's memory, thereby enhancing the user's conversational experience. To address this, we propose using a decay ratio $\eta_{decay}$ less than 1. After each round of dialogue, the stored entropy cache $E$ is multiplied by this decay ratio $E=E \times \eta_{decay}$, allowing the model to naturally forget older key information and focus more on recent critical information. The overall process of SirLLM can be referred to in Algorithm \ref{alg:sirllm}.

\begin{algorithm} 
	\caption{Streaming Infinite Retentive LLM} 
	\label{alg:sirllm} 
 	\renewcommand{\algorithmicrequire}{\textbf{Input:}}
	\renewcommand{\algorithmicensure}{\textbf{Output:}}

	\begin{algorithmic}
 \REQUIRE i-th turn's user input $I_i=\{x_1,x_2,...,x_n\}$
 \ENSURE  i-th turn's system response $R_i=\{r_1,r_2,...,r_m\}$
 % \IF{KV cache size $\leq$ L}
 % \STATE{$R_i=LLM(cache,I_i)$}
 % \STATE{$E_i=Entropy([I_i,R_i])=\{e_1,e_2,$\\
 % $...,e_{n+m}\}$}
 % \STATE{$E \leftarrow E+E_i$}
 % \STATE{$ E=E \times decay$}
 \FOR{turn $t$ in range(i)}
 \IF{KV cache size \textgreater L}
 \STATE{$\mathrm{Id_{entropy}}=\mathrm{Top_{k}(E)}$}
 \STATE{$\mathrm{KV}_{cache} \leftarrow  \mathrm{KV}_{cache}[\mathrm{Id_{sink},Id_{entropy}}]$}
  \STATE{$E \leftarrow E[\mathrm{Id_{sink},Id_{entropy}}]$}
  \ENDIF
  \STATE{$R_t,\mathrm{KV}_{cache}=LLM(\mathrm{KV}_{cache},I_t)$}
 \STATE{$E_t=Entropy([I_t,R_t])=\{e_1,e_2,$\\
 $...,e_{n+m}\}$}
 \STATE{$E \leftarrow E+E_t$}
 \STATE{$ E=E \times \eta_{decay}$}
\ENDFOR
	\end{algorithmic} 
\end{algorithm}

\section{Experiments}
\subsection{Experimental Setup}
%介绍一下evaluate的计算方法：将每个选项拼接在之后，然后看那个选项的logit最大
%超参数设置

We tested SirLLM on two different categories of large models: Vicuna-13b-v1.3, Vicuna-7b-v1.3 \cite{DBLP:journals/corr/abs-2306-05685}, Yi-34B-Chat, Yi-6B-Chat \footnote{https://github.com/01-ai/Yi}. %\cite{Yi}.
Following the evaluation methodologies used in \cite{DBLP:conf/nips/BrownMRSKDNSSAA20,DBLP:journals/corr/abs-2307-09288,eval-harness}, we evaluate the performance of SirLLM on various datasets by appending different option letters to the answers. We then calculate the logits for each option and select the option with the highest logits as the final answer. All experiments were conducted on an NVIDIA A800 GPU.

\subsubsection{Baslines}
To comprehensively evaluate the effectiveness of SirLLM, we utilized three baseline models:

\paragraph{StreamLLM:} StreamLLM \cite{xiao2023efficient} preserves the key-value states of only the attention sink tokens and recent tokens.

\paragraph{RandomLLM:} RandomLLM maintains the key-value states of the attention sink tokens as well as a random selection of historical tokens.

\paragraph{IntervalLLM:} Taking inspiration from \cite{DBLP:journals/corr/abs-1904-10509}, we developed IntervalLLM. This model, in addition to preserving attention sink tokens, uniformly samples tokens from the historical token sequence at fixed intervals. These intervals are adaptively determined, The size of these intervals is adaptively determined, calculated as $\text{interval}= \lfloor {\text{history token length} \over \text{cache size}} \rfloor$. This approach continues until the cache size is fully utilized

To ensure a fair comparison, in line with StreamLLM, all models preserve the KV cache states of attention sink tokens with a uniform size of 4 and we report the average accuracy for RandomLLM
\subsection{Results}
To thoroughly validate the effectiveness of the SirLLM framework, we designed three distinct tasks, each assessing SirLLM from a different perspective: (1) DailyDialog: This task evaluates SirLLM's conversational coherence and memory capabilities in everyday multi-turn dialogue scenarios. (2) Grocery Shopping: In this task, we focus on assessing SirLLM's memory capabilities. Initially, the LLM is informed about the groceries to be purchased. Subsequent rounds of commonsense QA with the LLM are conducted, culminating in a query to ascertain if SirLLM remembers the required groceries. (3) Rock-Paper-Scissors: In this task, by engaging in multiple rounds of rock-paper-scissors with users having distinct throwing preferences, we test whether SirLLM can utilize its enhanced memory ability to analyze historical information, discern users' throwing preferences, and thereby maximize its winning probability. 
\subsubsection{DailyDialog}
\paragraph{Dataset Construction} To assess the performance of SirLLM in everyday dialogue scenarios, we evaluate SirLLM using the DailyDialog dataset \cite{DBLP:conf/ijcnlp/LiSSLCN17}. DailyDialog is a high-quality, multi-turn, open-domain English dialogue dataset. To measure SirLLM's effectiveness more intuitively, we have reformatted DailyDialog into a multiple-choice question format, where SirLLM is tasked with selecting the most appropriate response. We have selected a sample from the constructed DailyDialog dataset, as illustrated in Figure \ref{fig:dailydialog-sample} in Appendix \ref{append:samples}. 
%More specifically, we modified the test split of the DailyDialog dataset to create a set of four-option multiple-choice questions. This set includes one correct option and three dummy choices, which are selected from the validation split. 
For more detailed statistics and construction process about the modified dataset, please refer to Table \ref{tab:dailydialog_statistic} in Appendix \ref{append:sta}. From the Table \ref{tab:dailydialog_statistic}, we observe that the average number of tokens per turn in the modified DailyDialog dataset is approximately 461.55. Therefore, we have set the cache size to 512. It was found that 199 dialogs in the dataset have token counts exceeding 512. In such longer dialog scenarios, SirLLM can be highly effective. By enabling the LLM to remember only key tokens, SirLLM is endowed with a longer memory span. This capability allows it to engage more effectively in extended dialogues.

\paragraph{Results}

\begin{table}[h]
\centering
\scalebox{0.75}{
\begin{tabular}{lcccccc}
\toprule
                 & \# Entropy & \# Recent & $\eta_{decay}$ & ACC(\%)            & $\Delta$         \\ \midrule
\multicolumn{6}{l}{\textit{\textbf{Yi-6b} Attention Sink Size: 4} }      \\
Stream                         & 0              & 508         & 1           & 76.90          &               \\
Random                         & 508            & 0           & 1           & 71.10          & -5.80         \\
 Interval                          & 508            & 0           & 1           & 65.20          & -11.70       \\ \midrule
\textbf{SirLLM}                         & 508            & 0           & 0.7         & \textbf{83.85} & \textbf{6.95} \\ \midrule
\multicolumn{6}{l}{\textit{\textbf{Yi-34b} Attention Sink Size: 4}}          \\
Stream                         & 0              & 508         & 1           & 85.35          &               \\
Random                         & 508            & 0           & 1           & 82.17          & -3.18         \\
 Interval                          & 508            & 0           & 1           & 70.70          & -14.65        \\ \midrule
\textbf{SirLLM}                         & 508            & 0           & 0.7         & \textbf{90.35} & \textbf{5.00} \\ \midrule
\multicolumn{6}{l}{\textit{\textbf{Vicuna-7b}  Attention Sink Size: 4}}            \\
Stream                         & 0              & 508         & 1           & 57.55          &               \\
Random                         & 508            & 0           & 1           & 57.48 & -0.13  \\
 Interval                          & 508            & 0           & 1           & 54.45          &  -3.10        \\ \midrule
\textbf{SirLLM }                        & 508            & 0           & 0.5         & \textbf{59.15}&          \textbf{1.60}         \\ \midrule
\multicolumn{6}{l}{\textit{\textbf{Vicuna-13b} Attention Sink Size: 4}}         \\
Stream                         & 0              & 508         & 1           & 71.10          &               \\
Random                         & 508            & 0           & 1           & 69.27          & -1.83         \\
 Interval                          & 508            & 0           & 1           & 62.05          & -9.05         \\ \midrule
\textbf{SirLLM}                         & 508            & 0           & 0.6         & \textbf{71.40} & \textbf{0.30} \\ \bottomrule
\end{tabular}
}
\caption{Results for the DailyDialog dataset are presented as follows: \# Entropy and \# Recent indicate the cache sizes allocated for tokens with the highest entropy and for recent tokens, respectively. ACC (\%) represents the accuracy. $\Delta$ signifies the improvement of the model relative to the baseline StreamLLM.}
\label{tab:dailydialog_result}
\end{table}

In the table \ref{tab:dailydialog_result}, to ensure a fair comparison, each model is configured with a unified KV cache size of 512.
Table \ref{tab:dailydialog_result} displays the performance of various models on the DailyDialog dataset. It is evident that SirLLM demonstrates a clear advantage over three baseline models across four different LLMs. It is noteworthy that SirLLM's performance remains consistently stable, whereas RandomLLM and IntervalLLM sometimes even lead to performance degradation. When employing Yi-34b, SirLLM achieved the highest accuracy of 90.35\% on the modified DailyDialog dataset, marking an impressive 5.00\% increase in accuracy compared to StreamLLM. These results robustly demonstrate SirLLM's capability to enhance the memory ability of LLMs, providing them with a longer retention span and thereby offering users a smoother conversational experience.

\subsubsection{Grocery Shopping}
\paragraph{Dataset Construction} To more vividly demonstrate SirLLM's superior memory capabilities, we designed the second task, Grocery Shopping, based on the CommonsenseQA (CSQA) \cite{DBLP:conf/naacl/TalmorHLB19} dataset to create the Grocery Shopping dataset. Specifically, in the first interaction, the user informs the LLM of the groceries they wish to purchase. This is followed by 20 rounds of commonsense QA sessions with the LLM, where the questions are sourced from the train and development splits of the CSQA dataset and formatted as multiple-choice questions. After these 20 rounds, the user then asks the LLM to recall and select the required groceries from four options. This task is designed to test the LLM's long-term memory through the grocery-related questions and its ability to maintain excellent short-term memory and reasoning skills through the commonsense QA. The detailed dataset statistics can be found in Table \ref{tab:grocery_statistic} in Appendix \ref{append:sta} and dataset samples can be found in Figure \ref{fig:grocery-sample} in Appendix \ref{append:samples}. From the table, we can see that the average token length per dialog is 1223.81 and all the 548 dialogs' total token number exceeds 1024. Therefore, we set the cache size for Grocery Shopping as 1024.
\paragraph{Result}
In the Grocery Shopping task, to enable the model to maintain a longer memory, we uniformly set the decay ratio to 1. The overall results can be found in Table \ref{tab:grocery_shopping_result}. 
% Overall, SirLLM's performance validates its advantages as an optimized LLM variant in sustaining long-term memory, especially in applications that require retaining extensive historical information for effective dialogue.

\begin{table}[h]
\centering
\scalebox{0.7}{
\begin{tabular}{lllllcc}
\toprule
       & \# Entropy & \# Recent & $\text{ACC}_{c}$     & $\text{ACC}_{g}$  & $\Delta_{c}$  & $\Delta_{g}$ \\ \midrule
\multicolumn{7}{l}{\textit{\textbf{Yi-6b}} Attention Sink Size: 4; $\eta_{decay}=1$}                                                              \\
Stream & 0              & 1020        & 71.33          & 25.73          &               &                \\
Random & 1020           & 0           & 70.33          & 77.55          & -1.00         & 51.82          \\
 Interval  & 1020           & 0           & 63.98          & 21.72          & -7.20         & -4.01          \\ \midrule
 \textbf{SirLLM} & 1020           & 0           & \textbf{72.44} & \textbf{99.27} & \textbf{1.11} & \textbf{73.54}

  \\ \midrule
\multicolumn{7}{l}{\textit{\textbf{Vicuna-7b}} Attention Sink Size: 4; $\eta_{decay}=1$}                                                          \\
Stream & 0              & 1020        & 50.84          & 28.65          &               &                \\
Random & 1020           & 0           & 50.97 & 85.04          & 0.13 & 56.39          \\
 Interval  & 1020           & 0           & 47.21          & 23.72          & -3.63         & -4.93          \\\midrule
\textbf{SirLLM} & 1020           & 0           & \textbf{51.04}          & \textbf{96.17} & \textbf{0.20}          & \textbf{67.52} \\ \midrule
\multicolumn{7}{l}{\textit{\textbf{Vicuna-13b}} Attention Sink Size: 4; $\eta_{decay}=1$}                                                         \\
Stream & 0              & 1020        & 60.10          & 24.45          &               &                \\%\midrule
%Random & 1020           & 0           & 59.97          & 81.75          & -0.13         & 57.30          \\
% Interval  & 1020           & 0           & 54.30          & 21.35          & -5.80          & -3.10          \\\midrule
\textbf{SirLLM} & 1020           & 0           & \textbf{60.23} & \textbf{97.08} & \textbf{0.13} & \textbf{72.63} \\ \midrule

\multicolumn{7}{l}{\textit{\textbf{Yi-34b}} Attention Sink Size: 4; $\eta_{decay}=1$}                                                             \\
Stream & 0              & 1020        & 81.35          & 26.29          &               &                \\%\midrule
%Random & 1020           & 0           & 80.95          & \textbf{96.41} & -0.40         & \textbf{70.12} \\
 %Interval  & 1020           & 0           & 72.51          & 21.72          & -8.84         & -4.57          \\ \midrule
 \textbf{SirLLM} & 1020           & 0           & \textbf{81.44} & \textbf{89.60}          & \textbf{0.09} & \textbf{63.31}         \\ 
 \bottomrule
\end{tabular}
}
\caption{Results for the Grocery Shopping dataset: \# Entropy and \# Recent indicate the cache sizes allocated for tokens with the highest entropy and for recent tokens, respectively. $\text{ACC}_{c}$ and $\text{ACC}_{g}$ represents the accuracy for commonsense QA and Grocery Shopping, respectively. $\Delta_{c}$ and $\Delta_{g}$ signify the improvement of the model relative to the baseline StreamLLM.
}
\label{tab:grocery_shopping_result}
\end{table}

Table \ref{tab:grocery_shopping_result} clearly indicates that SirLLM consistently demonstrates an improvement in accuracy across different models. Specifically, SirLLM not only maintains its commonsense question-answering abilities that require short-term memory but also shows a substantial enhancement in memory capabilities for the Grocery Shopping task. This outcome is attributed to SirLLM's effective utilization of larger cache space allocated for key information, allowing it to maintain more contextual information in extended dialogues. This underscores SirLLM's efficacy not only in specific tasks but also in maintaining its memory advantage across different types of tasks, which is crucial for building a more adaptable and multifunctional dialogue system.

\begin{table*}[h]
\centering
\scalebox{0.78}{
\begin{tabular}{llllcccccccccc}
\toprule
          & \multicolumn{1}{c}{\multirow{2}{*}{\# Entropy}} & \multicolumn{1}{c}{\multirow{2}{*}{\# Recent}} & \multicolumn{1}{c}{\multirow{2}{*}{$\eta_{decay}$}} & \multicolumn{3}{c}{Paper} & \multicolumn{3}{c}{Rock} & \multicolumn{3}{c}{Scissors} & \multicolumn{1}{c}{Average}                                                  \\ \cline{5-14} 
          & \multicolumn{1}{c}{}                            & \multicolumn{1}{c}{}                           & \multicolumn{1}{c}{}                       & win     & tie    & lose   & win    & tie    & lose   & win      & tie     & lose    & win                               \\ \midrule
\multicolumn{14}{l}{\textit{Yi-6b}}                                                                                                                                                                                                                                                                                                 \\
Stream & 0                                               & 1020                                           & 1                                          & 31.10   & 19.60  & 49.30  & 30.90  & 19.10  & 50.00  & 46.45    & 31.00   & 22.55   & 36.15                        \\
Random    & 1020                                            & 0                                              & 1                                          & 20.00   & 49.45  & 30.55  & 49.73  & 31.02  & 19.25  & 27.18    & 21.93   & 50.88   & 32.31                     \\
 Interval     & 1020                                            & 0                                              & 1                                          & 19.45   & 50.15  & 30.40  & 50.00  & 30.90  & 19.10  & 27.35    & 20.8   & 51.85   & 32.27            \\ \midrule
\textbf{SirLLM}    & 1020                                            & 0                                              & 0.9         & 30.65   & 19.55  & 49.80  & 30.90  & 19.10  & 50.00  & 50.45    & 28.05   & 21.50   & \textbf{37.33}                 \\ \hline
\multicolumn{14}{l}{\textit{Yi-34b}}                                                                                                                                                                                                                                                                                                \\
Stream & 0                                               & 1020                                           & 1                                           & 48.55   & 27.95  & 23.50  & 41.05  & 26.15  & 32.80  & 32.20    & 38.90   & 28.90   & 40.60  \\
Random    & 1020                                            & 0                                              & 1                                          & 30.57   & 19.68  & 49.62  & 19.08  & 50.12  & 30.97  & 51.70    & 27.35   & 20.95   & 33.78                     \\
 Interval     & 1020                                            & 0                                              & 1                                          & 45.40   & 26.05  & 28.55  & 35.45  & 24.30  & 40.25  & 26.20    & 46.15   & 27.65   & 35.68                    \\ \midrule
\textbf{SirLLM}    & 1020                                            & 0                                              & 0.8         & 48.45   & 28.00  & 23.55  & 40.60  & 26.15  & 33.25  & 37.05    & 38.25   & 24.70   & \textbf{42.03}            \\ \hline
\multicolumn{14}{l}{\textit{Vicuna-7b}}                                                                                                                                                                                                                                                                                             \\
Stream & 0                                               & 1020                                           & 1           & 28.75   & 24.05  & 47.20  & 19.80  & 45.90  & 34.30  & 51.20    & 27.70   & 21.10   & 33.25                     \\
Random    & 1020                                            & 0                                              & 1                                          & 19.57   & 49.62  & 30.82  & 49.82  & 31.03  & 19.15  & 27.68    & 20.90   & 51.42   & 32.36                     \\
 Interval     & 1020                                            & 0                                              & 1                                          & 29.15   & 37.95  & 32.90  & 24.40  & 45.80  & 29.80  & 28.25    & 35.05   & 36.70   & 27.27                     \\ \midrule
\textbf{SirLLM}    & 1020                                            & 0                                              & 0.7         & 26.20   & 32.10  & 41.70  & 27.40  & 30.45  & 42.15  & 48.60    & 29.85   & 21.55   & \textbf{34.07 }           \\ \hline
\multicolumn{14}{l}{\textit{Vicuna-13b}}                                                                                                                                                                                                                                                                                            \\
Stream & 0                                               & 1020                                           & 1                                          & 30.25   & 21.15  & 48.60  & 22.60  & 47.25  & 30.15  & 51.20    & 27.65   & 21.15   & 34.68                    \\
Random    & 1020                                            & 0                                              & 1                                          & 44.02   & 26.63  & 29.35  & 30.43  & 21.70  & 47.97  & 28.32    & 46.43   & 25.25   & 34.26                     \\
 Interval     & 1020                                            & 0                                              & 1                                          & 29.80   & 22.65  & 47.55  & 21.80  & 45.65  & 32.50  & 50.25    & 27.80   & 21.90   & 33.95                     \\ \midrule
\textbf{SirLLM}    & 1020                                            & 0                                              & 0.7         & 28.50   & 26.20  & 45.30  & 33.70  & 39.80  & 26.50  & 48.10    & 25.75   & 26.15   & \textbf{36.77 }          \\ \hline
\end{tabular}
}
\caption{
%Results for the Rock-Paper-Scissors dataset are presented as follows: \# Entropy and \# Recent indicate the cache sizes allocated for tokens with the highest entropy and for recent tokens, respectively. Rock, Paper and Scissors represents player with corresponding throwing preferences. Win, Tie and Lose stands for win rate(\%),tie rate(\%) and lose rate (\%), respectively.
Results for the Rock-Paper-Scissors dataset. \# Entropy and \# Recent denote the allocated cache sizes for tokens with the highest entropy and for the most recent tokens, respectively. 'Rock,' 'Paper,' and 'Scissors' correspond to players with a preference for each respective move. 'Win,' 'Tie,' and 'Lose' represent the win rate (\%), tie rate (\%), and loss rate (\%), respectively.
}
\label{tab:rps_result}
\end{table*}

\subsubsection{Rock-Paper-Scissors}

\paragraph{Dataset Construction} 
To better observe the performance of SirLLM in scenarios with infinitely long streaming dialogue inputs, we constructed a Rock-Paper-Scissors dataset. In this dataset, we created three players with preferences for throwing rock, paper, or scissors, respectively. In each round, we inform the LLM of the previous round's user move and the outcome, and then we ask the LLM to analyze the user's throwing preferences to maximize its own winning rate for the next round. Detailed information about the dataset and the probabilities of each player's moves can be found in Table \ref{tab:rps_statistic} in Appendix \ref{append:sta}. A sample of the data is illustrated in Figure \ref{fig:rps-sample} in Appendix \ref{append:samples}. Unlike the DailyDialog and Grocery Shopping datasets, where the KV cache is reset to zero after each round, the Rock-Paper-Scissors task allows the LLM to engage in 2000 rounds of play without resetting the KV cache, achieving a truly infinite number of dialogue turns. This aims to observe whether SirLLM can remember key information and more user historical preferences to better maximize its win rate.

\paragraph{Result}
The results showcased in Table \ref{tab:rps_result} for the Rock-Paper-Scissors dataset reveal that SirLLM consistently surpasses the baseline StreamLLM for players with varied throwing preferences. Upon closer examination of the data, it becomes apparent that SirLLM delivers a steady enhancement in win rates against players of different preferences, maintaining this enhanced performance uniformly across all the models evaluated. 
Furthermore, the decay mechanism integrated within SirLLM plays a crucial role in sustaining a balanced performance over numerous rounds, as reflected by its uniformly elevated win rates. 
This characteristic of SirLLM proves especially advantageous in scenarios involving extended interactions, such as long-duration Rock-Paper-Scissors games, where the model's capacity to adapt and recall previous moves is imperative for success.

\begin{table}[h]
\centering
\scalebox{0.8}{
\begin{tabular}{lrrrr}
\toprule
      & $\text{ACC}_{c}$     & $\text{ACC}_{g}$  & $\Delta_{c}$  & $\Delta_{g}$ \\ \midrule
\multicolumn{5}{l}{\textit{Yi-6b}}                               \\
Stream & 71.33     & 25.73        &             &                \\
1-shot & 58.66     & 25.00        & -12.67      & -0.73          \\
2-shot & 63.95     & 25.36        & -7.38        & -0.37           \\
3-shot & 65.42     & 23.72        & -5.91        & -2.01           \\\midrule
\textbf{SirLLM} & 72.44     & 99.27        & 1.11        & \textbf{73.54}          \\ \midrule
\multicolumn{5}{l}{\textit{Yi-34b}}                              \\
Stream & 81.35     & 26.29        &             &                \\
1-shot & 75.14     & 23.91        & -6.21       & -2.38          \\
2-shot & 78.50     & 24.64        & -2.85       & -1.65          \\
3-shot & 79.20     & 25.18        & -2.15       & -1.11          \\\midrule
\textbf{SirLLM} & 81.44     & 89.60        & 0.09        & \textbf{63.31}          \\ \midrule
\multicolumn{5}{l}{\textit{Vicuna-7b}}                           \\
Stream & 50.84     & 28.65        &             &                \\
1-shot & 48.54     & 27.01        & -2.30       & -1.64          \\
2-shot & 49.11     & 27.19        & -1.73       & -1.46          \\
3-shot & 49.81     & 27.55        & -1.03       & -1.10          \\\midrule
\textbf{SirLLM} & 51.04     & 96.17        & 0.20        & \textbf{67.52}          \\ \midrule
\multicolumn{5}{l}{\textit{Vicuna-13b}}                          \\
Stream & 60.10     & 24.45        &             &                \\
1-shot & 55.34     & 22.26        & -4.76       & -2.19          \\
2-shot & 58.94     & 26.46        & -1.16       & 2.01           \\
3-shot & 60.44     & 27.01        & 0.34        & 2.56          \\ \midrule
\textbf{SirLLM} & 60.23     & 97.08        & 0.13        & \textbf{72.63}          \\ \bottomrule
\end{tabular}}
\caption{Few-shot results for Grocery Shopping dataset}
\label{tab:few-shot}
\end{table}

\section{Further Exploration}
\subsection{Few-shot}
\begin{figure*}[h]
    \centering
    \includegraphics[width=1\linewidth]{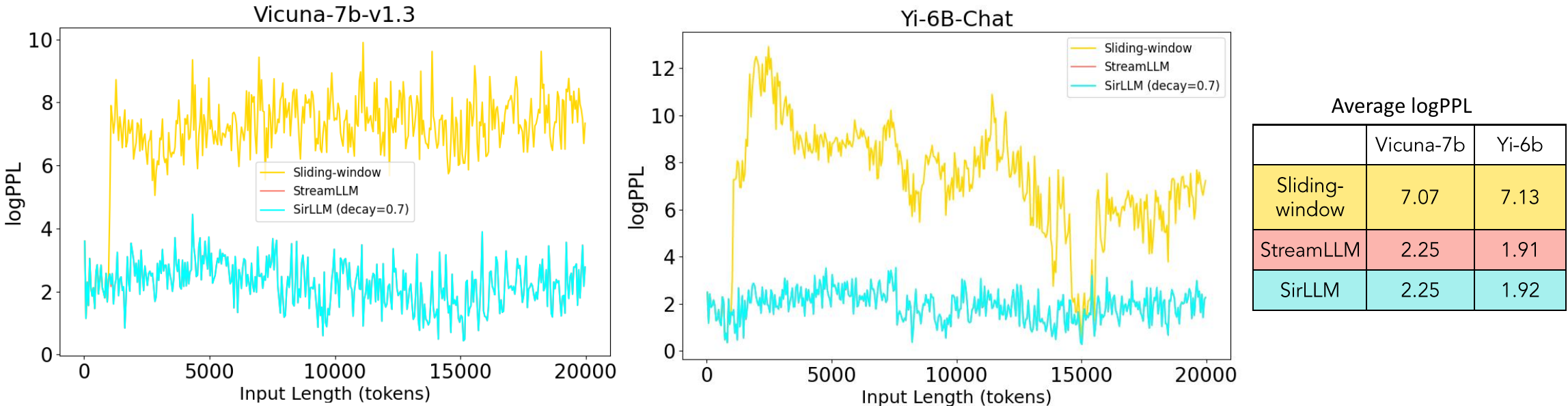}
    \caption{The perplexity of language modeling on 20K token text. The Sliding-window's PPL escalates dramatically once the token length exceeds the pre-trained length. In contrast, both SirLLM and StreamLLM, which incorporate attention sink tokens, show stable performance. SirLLM and StreamLLM's performances are almost identical, effectively demonstrating that SirLLM's memory mechanism does not impair the model's answering performance and can indeed reinforce the model's memory capabilities.}
    \label{fig:PPL}
\end{figure*}

\citet{DBLP:conf/nips/BrownMRSKDNSSAA20} demonstrates that few-shot learning can significantly aid models in reasoning and answering questions. SirLLM, by eliminating redundant KV cache, achieves enhanced memory capabilities, which translates into improved performance on the CSQA dataset. This improvement could also be interpreted as SirLLM's ability to incorporate more few-shot exemplars with less cache, thereby attaining higher accuracy. On this premise, we compared SirLLM with 1-shot, 2-shot, and 3-shot learning approaches, with results as presented in Table \ref{tab:few-shot}. In n-shot experiments, we prepend the preceding n questions as few-shot exemplars before each question, aiming to simulate an input format similar to that of StreamLLM. As shown in the table, SirLLM not only improves upon the baseline StreamLLM in both the CSQA and Grocery Shopping datasets, but it also maintains this enhanced performance despite the increment in the number of shots. This consistency underscores the model's ability to leverage the rich information contained within the few-shot examples without becoming overwhelmed by the increased data.

%effectively use context provided by few-shot exemplars, which is a testament to its robust reasoning and memory mechanisms. Notably, even when additional context is provided in the form of more shots, SirLLM's performance remains stable or improves, suggesting that its architecture is particularly suited for leveraging the rich information contained within the few-shot examples without becoming overwhelmed by the increased data.

\subsection{PPL for long text}
%和streamllm一样绘制Language modeling perplexity on texts with 20K tokens across various LLM【figure 3】

Following the approach of StreamLLM, we plotted the log Perplexity (logPPL) of SirLLM, StreamLLM, and Sliding-window on texts spanning 20,000 tokens across various LLMs, as depicted in the Figure \ref{fig:PPL}. The Figure reveals that while the Sliding-window model exhibits volatility in PPL, particularly beyond the length it was trained on, SirLLM maintains a consistent and stable PPL, suggesting a robustness to input length. The average logPPL values in the accompanying table further corroborate this, with SirLLM matching StreamLLM's performance closely across both Vicuna-7b and Yi-6b models. This indicates that SirLLM and StreamLLM have comparable short-term memory capabilities, with SirLLM not adversely affecting the model's ability to retain information over shorter durations.  This alignment of PPL between SirLLM and StreamLLM, despite SirLLM's enhanced memory function, underscores the efficacy of SirLLM's design in managing longer context without compromising the language model's fluency or coherence.

\section{Conclusion}
Addressing the critical challenges of managing infinite input lengths and maintaining memory capability, SirLLM harmonizes long dialogue retention without the necessity of model fine-tuning by selectively fortifies the model's focus on pivotal information. Through experiments across three tailor-made tasks: DailyDialog, Grocery Shopping, and Rock-Paper-Scissors, SirLLM has demonstrated a consistent and stable improvement over existing models, irrespective of the complexity and length of the dialogues. The experimental outcomes validate the robustness and versatility of SirLLM, making it an invaluable asset for future explorations and applications in natural language processing.

% Bibliography entries for the entire Anthology, followed by custom entries
%\bibliography{anthology,custom}
% Custom bibliography entries only
\section*{Limitation}
The limitations of SirLLM include: (1) Adaptation to Various Scenarios: Currently, users may need to manually adjust the decay ratio to achieve desired outcomes in different application scenarios. Developing an adaptive mechanism that automatically tunes the decay ratio based on specific contexts presents a viable direction for future work. (2) Significance Discrepancy: What users consider important information may not always align with the model's criteria, leading to potential omissions in memory retention. Therefore, a more accurate mechanism for cache retrieval and storage warrants detailed exploration in future research endeavors. This could ensure that the model better aligns with user priorities and improves overall recall accuracy.

\bibliography{custom}

\appendix

\newpage
% \section{Original Work Attribution}

\section{Dataset Statistics}
\label{append:sta}
\subsection{DailyDialog}

\begin{table}[h]
\centering
\begin{tabular}{lr}
\toprule
DailyDialog             & \multicolumn{1}{l}{Statistics}   \\ \midrule
\#dialogs               & 518                          \\
\#average turn          & 3.85                         \\
\#average token (dialog)  & 461.55                         \\
\#average word (dialog)   & 309.92                        \\
\# dialogs ($\geq 512$) & 199                           \\ \bottomrule
\end{tabular}
\caption{Detailed statistics of DailyDialog(modified)}
\label{tab:dailydialog_statistic}
\end{table}
% Table \ref{tab:dailydialog_statistic} illustrates the detailed statistics of the modified DailyDialog dataset where \#dialogs denotes the number of dialogs, \#average turn denotes the average number of turns per dialog, \#average token (turn) means the average number of tokens per turn using Vicuna-7b-v1.3 as tokenizer, \#average word (turn) denotes the average number of words per turn, and \# dialogs ($\geq 512$) denotes the count of dialogs where the number of tokens exceeds 512.
We modified the test split of the DailyDialog dataset to create a set of four-option multiple-choice questions. This set includes one correct option and three dummy choices, which are selected from the validation split. Table \ref{tab:dailydialog_statistic} presents the detailed statistics of the modified DailyDialog dataset. In this table, \#dialogs indicates the total number of dialogs; \#average turn refers to the average number of turns per dialog; \#average token (dialog) represents the average number of tokens per dialog, calculated using the Vicuna-7b-v1.3 tokenizer; \#average word (dialog) signifies the average number of words per dialog; and \#dialogs ($\geq 512$) shows the count of dialogs where the total number of tokens exceeds 512. 

\subsection{Grocery Shopping}
\begin{table}[h]
\centering
\begin{tabular}{lr}
\toprule
Grocery Shopping         & Statistics \\ \midrule
\#dialogs                & 548        \\
\#groceries              & 53           \\
\#average turn           & 22         \\
\#average token (dialog) & 1223.81    \\
\#average word (dialog)  & 631.60     \\
\# dialogs ($\geq 1024$)  & 548        \\ \bottomrule
\end{tabular}
\caption{Detailed statistics of Grocery Shopping}
\label{tab:grocery_statistic}
\end{table}

Table \ref{tab:grocery_statistic} presents the detailed statistics of the Grocery Shopping dataset. In this table, \#dialogs indicates the total number of dialogs; \#groceries represents the number of different types of groceries;  \#average turn refers to the average number of turns per dialog; \#average token (dialog) represents the average number of tokens per dialog, calculated using the Vicuna-7b-v1.3 tokenizer; \#average word (dialog) signifies the average number of words per dialog; and \#dialogs ($\geq 1024$) shows the count of dialogs where the total number of tokens exceeds 1024.

\subsection{Rock-Paper-Scissors dataset}

% Please add the following required packages to your document preamble:
% \usepackage{multirow}

\begin{table}[h]
\centering
\begin{tabular}{lll}
\hline
\multicolumn{2}{l}{Rock-Paper-Scissors}                                                   & Statistics \\ \hline
\multicolumn{2}{l}{\#rounds}                                                              & 2000       \\
\multicolumn{2}{l}{\#average token (rounds)}                                              & 54         \\
\multicolumn{2}{l}{\#average word (rounds)}                                               & 35         \\ \hline
\multirow{3}{*}{\begin{tabular}[c]{@{}l@{}}Player 1\\ (Rock)\end{tabular}}     & rock     & 0.5        \\
                                                                               & paper    & 0.3        \\
                                                                               & scissors & 0.2        \\ \hline
\multirow{3}{*}{\begin{tabular}[c]{@{}l@{}}Player 2\\ (Paper)\end{tabular}}    & rock     & 0.2        \\
                                                                               & paper    & 0.5        \\
                                                                               & scissors & 0.3        \\ \hline
\multirow{3}{*}{\begin{tabular}[c]{@{}l@{}}Player 3\\ (Scissors)\end{tabular}} & rock     & 0.3        \\
                                                                               & paper    & 0.2        \\
                                                                               & scissors & 0.5        \\ \hline
\end{tabular}
\caption{Detailed statistics of Grocery Shopping}
\label{tab:rps_statistic}
\end{table}
Table \ref{tab:rps_statistic} presents the detailed statistics of the Rock-Paper-Scissors dataset. In this table, \#rounds indicates the total number of Rock-Paper-Scissors rounds; \#average token (rounds) represents the average number of tokens per rounds, calculated using the Vicuna-7b-v1.3 tokenizer; \#average word (rounds) signifies the average number of words per round. In the table \ref{tab:rps_statistic}, the preferences for each player's moves and their corresponding probabilities of throwing rock, paper, or scissors are also listed.

%不同的decay 对结果的影响
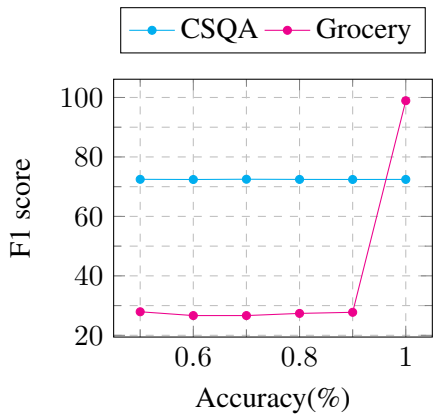
\begin{figure}[!h] %插入图片
\centering %图片居中

\begin{tikzpicture} %tikz图片

\begin{axis}[
    xlabel=Accuracy(\%), %横坐标名
    ylabel= F1 score, %纵坐标名
    height= 5 cm, 
    width=0.75 \linewidth,
    tick align=inside, %刻度在外显式
    legend columns=-1,
    legend style={at={(0.5,1.1)},anchor=south}, %图例在图下方显示
    minor tick num=1,
    grid=both,
    grid style=dashed
    ]

\addlegendentry{$\textrm{CSQA}$}
\addplot[color=cyan,mark=*,mark size=1.5pt,] coordinates{(0.5,72.47)(0.6,72.41)(0.7,72.5)(0.8,72.44)(0.9,72.42)(1,72.44)};
\addlegendentry{$\textrm{Grocery}$}
\addplot[color=magenta,mark=*,mark size=1.5pt,] coordinates{(0.5,27.92)(0.6,26.64)(0.7,26.64)(0.8,27.37)(0.9,27.74)(1,98.91)};

\end{axis}
\end{tikzpicture}
\caption{Performance of different decay ratio in Grocery Shopping dataset.}
\label{fig:decay_ratio}
\end{figure}

\section{The Impact of Decay Ratio on Memory Retention}
To more vividly illustrate the impact of the decay ratio on the memory capabilities of LLMs, we conducted experiments using various decay ratios in the Grocery Shopping task. The results of these experiments are presented in Figure \ref{fig:decay_ratio}. 
From the Figure \ref{fig:decay_ratio}, we can observe that when the decay ratio is set below one, the model completely forgets the groceries desired by the user after 20 rounds of commonsense question and answer sessions. However, adjusting the decay ratio does not significantly impact the model's performance on tasks requiring short-term memory, such as commonsense question answering. By fine-tuning the decay ratio, we can flexibly adapt the memory capabilities of the LLM to suit different scenarios. This effectively demonstrates the stability and efficacy of SirLLM's memory mechanism.

\section{Dataset Samples}
\label{append:samples}

\begin{figure}[h]
    \centering
    \includegraphics[width=1\linewidth]{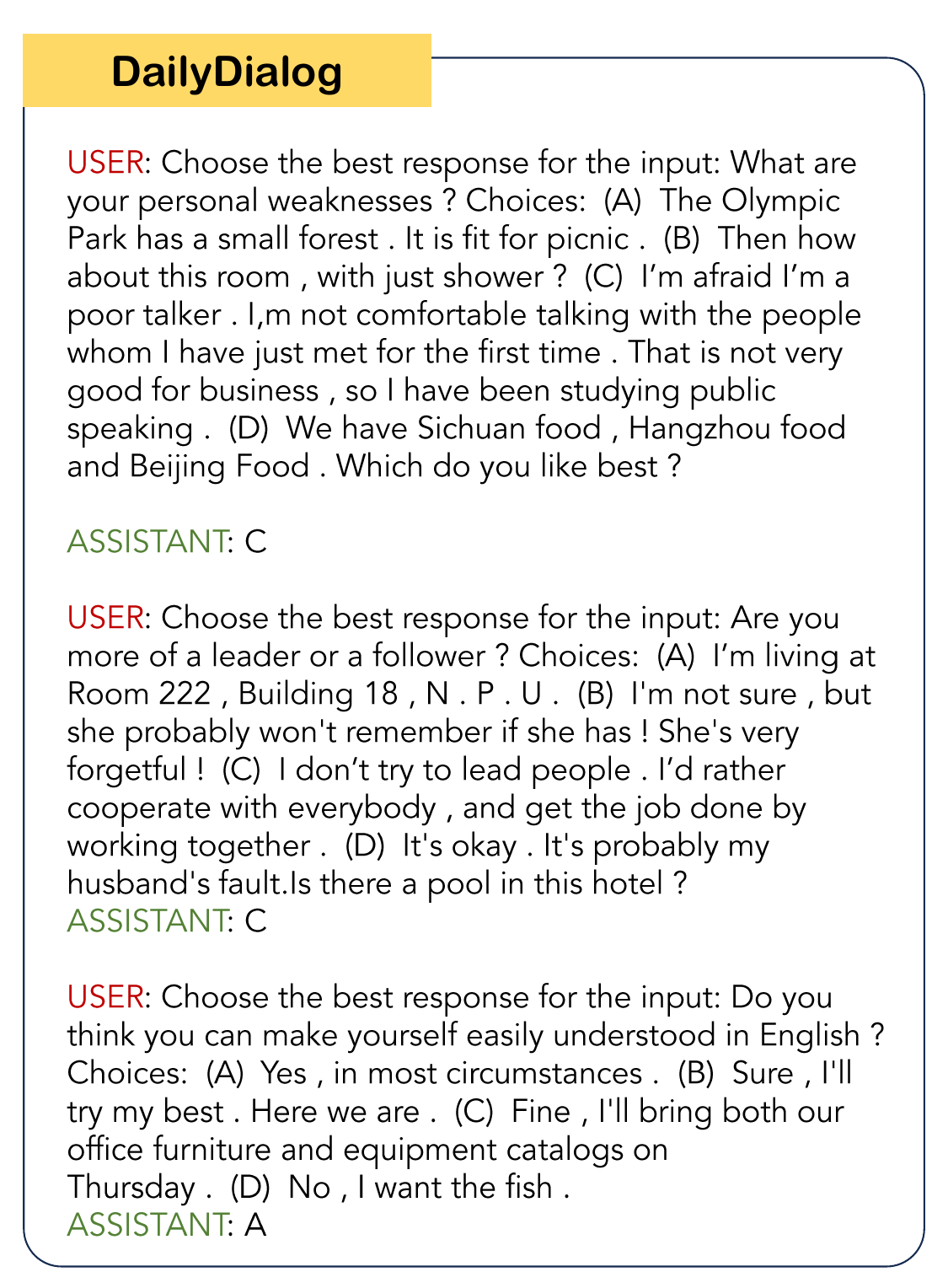}
    \caption{A sample from the DailyDialog dataset}
    \label{fig:dailydialog-sample}
\end{figure}

\begin{figure}[t]
    \centering
    \includegraphics[width=1\linewidth]{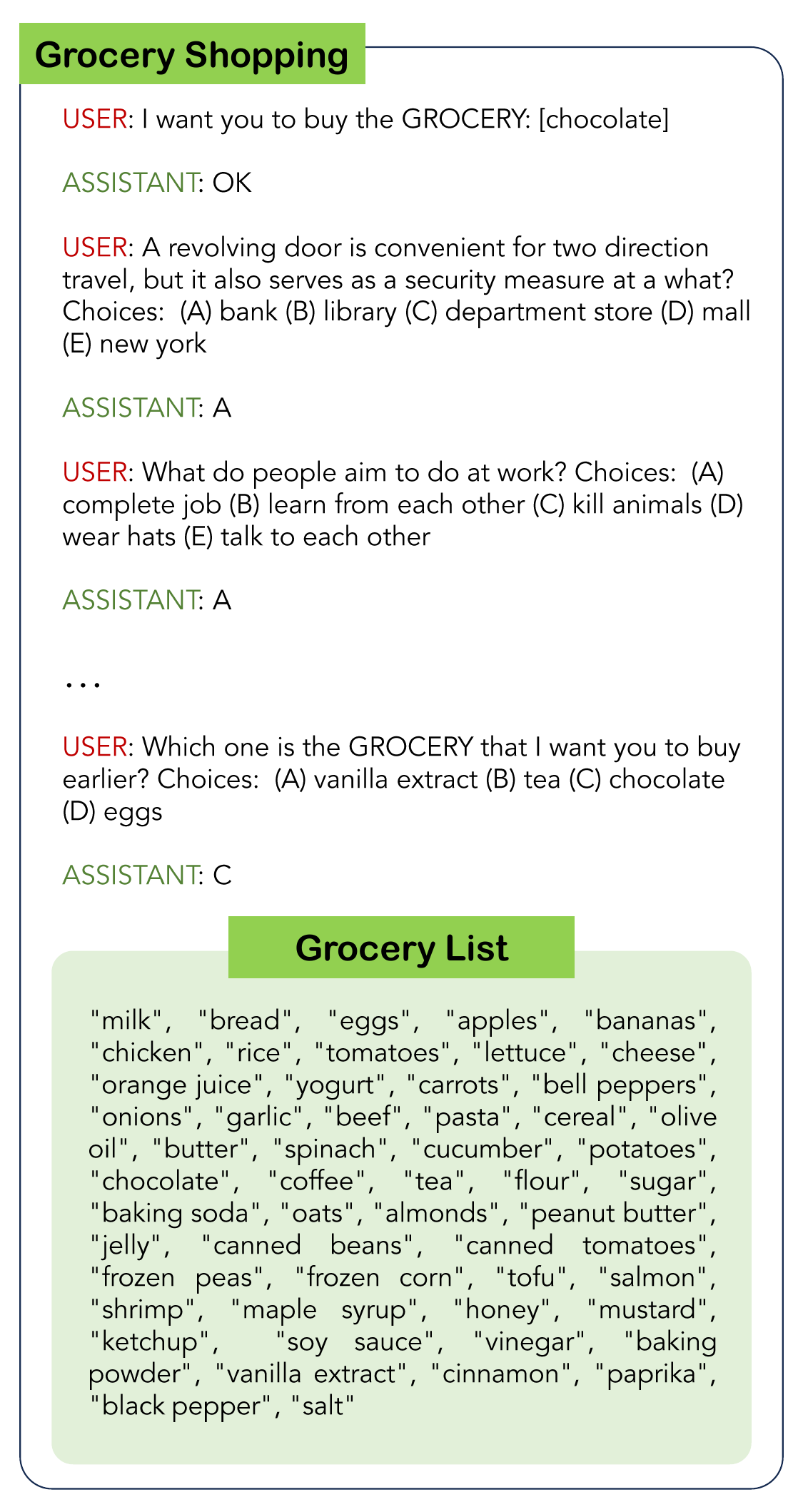}
    \caption{A sample from the Grocery Shopping dataset}
    \label{fig:grocery-sample}
\end{figure}

\begin{figure}
    \centering
    \includegraphics[width=1\linewidth]{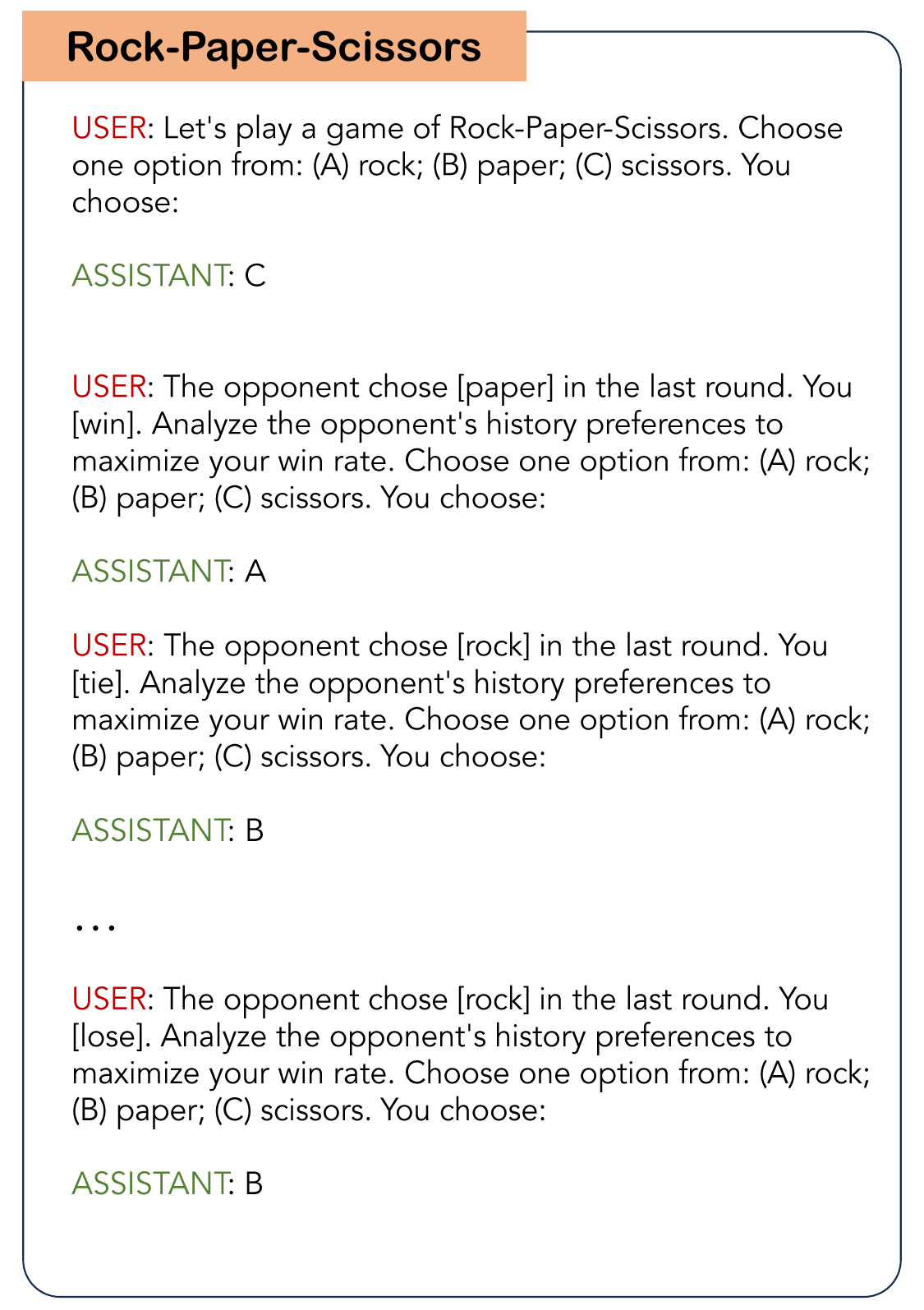}
    \caption{A sample from the Rock-Paper-Scissors dataset}
    \label{fig:rps-sample}
\end{figure}

\end{document}